\documentclass[journal]{IEEEtran}
\usepackage{amsmath,epsfig,esvect,mathtools,tabularx}
\DeclareMathOperator*{\argminA}{arg\,min}
\ifCLASSINFOpdf
\else
\fi
\begin{document}
%
\title{Using Sensory Time-cue to enable \\ Unsupervised Multimodal Meta-learning}
%
%
%

\author{Qiong~Liu
        and~Yanxia~Zhang
\thanks{Contact: qiongliu1 AT gmail.com}
\thanks{Manuscript received March 12, 2020.}}

%
%

\markboth{Journal of \LaTeX\ Class Files,~Vol.~14, No.~8, March~2020}%
{Shell \MakeLowercase{\textit{et al.}}: Using Sensory Time-cue for Unsupervised Meta-learning}
%



\maketitle

\begin{abstract}
As data from IoT (Internet of Things) sensors become ubiquitous, state-of-the-art machine learning algorithms face many challenges on directly using sensor data. To overcome these challenges, methods must be designed to learn directly from sensors without manual annotations. This paper introduces Sensory Time-cue for Unsupervised Meta-learning (STUM). Different from traditional learning approaches that either heavily depend on labels or on time-independent feature extraction assumptions, such as Gaussian distribution features, the STUM system uses time relation of inputs to guide the feature space formation within and across modalities. The fact that STUM learns from a variety of small tasks may put this method in the camp of Meta-Learning. Different from existing Meta-Learning approaches, STUM learning tasks are composed within and across multiple modalities based on time-cue co-exist with the IoT streaming data. In an audiovisual learning example, because consecutive visual frames usually comprise the same object, this approach provides a unique way to organize features from the same object together. The same method can also organize visual object features with the object's spoken-name features together if the spoken name is presented with the object at about the same time. This cross-modality feature organization may further help the organization of visual features that belong to similar objects but acquired at different location and time.  Promising results are achieved through evaluations.
\end{abstract}


%
\IEEEpeerreviewmaketitle

\section{Introduction}
%
%
%
%
\IEEEPARstart{S}{ince} AIoT (Artificial Intelligence and Internet of Things) becomes a new business trend, data from IoT systems is dramatically increasing. On the other hand, streaming all IoT sensor data to the cloud or saving data locally with big memory/disk space for machine learning may greatly increase cost and is not energy efficient for battery limited IoT devices. Because of above constraints and low energy consumption of sensors and computation units \cite{Pete2018}, the new business trend demands the capability of learning on-site instead of transmitting/saving raw data for offline learning. Learning directly from sensors on local devices may also enable efficient and real-time customization without using a large amount of unnecessary neural network weights.  Moreover, without the network bandwidth and disk space restrictions, a system is open to more training data to improve its performance.

Beyond network bandwidth and energy limitations, state-of-the-art supervised learning demands human experts to create annotations to guide training. Manual annotation is substantially slower than the speed of collecting data from sensors. With more and more sensors installed, it is difficult and expensive to annotate everything at any time by human. Moreover, people may not want to transfer all their personal data to the cloud shared with human labellers because of privacy concerns. In addition to above issues, existing interfaces for conducting supervised learning is also harder to use than the interface human use for teaching a baby. On the other hand, cutting-edge unsupervised learning approaches mostly use predefined time-independent assumptions, such as Gaussian distribution features, for `meaning' to surface in a feature space. However, perfect assumptions are difficult to find when data gets complicated. Because of these issues, there remains many challenges for learning from increased sensor data with existing learning approaches. To increase data handling speed, reduce data preparation and labelling cost, and enable data-driven meaning emergence, automating the learning process with direct sensor data is critical.

\begin{figure}[t]
\vskip 0.0in
\begin{center}
\centerline{\includegraphics[width=\columnwidth] {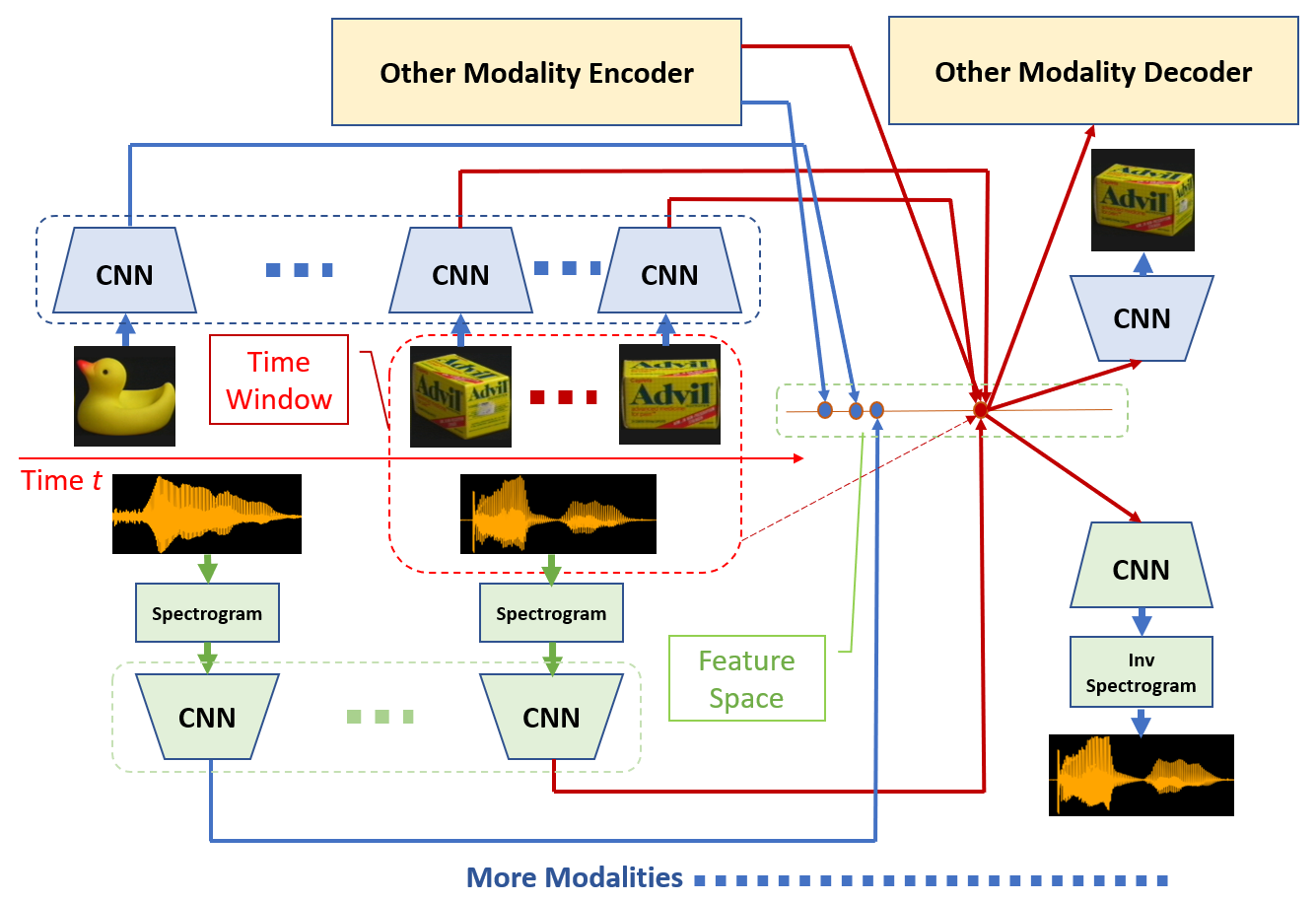}}
\caption{Architecture for time-cue-guided multi-modal sensor input learning.\label{fig:archfig}}
\end{center}
\vskip -0.4in
\end{figure}

Even though designing an algorithm to overcome all above issues sounds hard, those issues can easily be handled by natural biological systems that are `slower' than a powerful cloud. In this paper, we propose a learning approach motivated by biological systems. The main contribution of this paper is to use a time function to `supervise' the feature space formation and thus enable autonomous learning of sensor inputs. It also introduces different networks for different modalities in a traditional Siamese architecture. In the next section, we will explain the related work of the STUM model. Since there is no ready-to-use dataset, we use a section to explain how to prepare training and testing dataset. We then describe the details of the STUM model followed by an evaluation. We will end the paper with conclusions and discussions for future work.

\section{Related Works}
Since STUM learns from a variety of small tasks similar to testing tasks, we may consider it as a method in the camp of Meta-Learning \cite{Thrun1998}. Different from existing Meta-Learning approaches, such as Matching Networks \cite{Vinyals2016}, Prototypical Networks \cite{Snell2017}, or Relation Network \cite{Sung2018}, STUM learning tasks are composed within and across multiple modalities based on time-cue co-exist with the IoT streaming data. The STUM model has a shared feature space learning process similar to the Siamese architecture feature space formation process \cite{LeCun2006,Baldi1993}. Different from existing learning approaches based on the Siamese architecture, the STUM model uses a function defined over time for learning within and across modalities. More specifically, it uses time relation of inputs to form 'positive' groups and 'negative' groups in one or more modalities and use that info to influence the feature formation process. Additionally, STUM is not limited to twin identical networks for processing multi-modality data. Similar to the Structure from Motion (SfM) approach \cite{longuet1981,Hartley2004}, STUM also uses sequences of 2D images to form representations. Different from SfM which reconstructs accurate 3D representations with 2D sequences, STUM forms well-clustered representations in high dimensional feature space with 2D sequences.

There are also similarities between the STUM model and the triplet network \cite{Hoffer2015} or triplet learning \cite{Bengio2009} on learning data arrangement and weight sharing. Instead of using the triplet loss function or the triplet network loss function \cite{Hoffer2015, Bengio2009}, STUM uses a simple summation of contrastive loss as loss function for the model. With simple summation of contrastive loss, it is easier for a system to handle random number of data representations inside a time-window instead of waiting for data to form triplets before processing, which would be an issue for online learning. Compact representations trained with contrastive loss in the feature space also give the system more potential to learn complicated tasks. Contrastive loss and triplet loss usually need large-scale training data and the training may be slower than traditional classifiers. These issues may lead to big disadvantages if the data need to be prepared by human. On the other hand, if a machine can continuously get data from sensors and learn from these data automatically, these barriers will become much easier to overcome for many tasks.

Beyond relations with the Siamese Network and Triplet Network, training with the STUM model is related to traditional supervised learning, unsupervised learning, and reinforcement learning in the following ways. Similar to traditional supervised learning, STUM may involve two or more signals and can guide the learning process of one signal with another signal. Unlike traditional supervised learning, which explicitly specifies `supervisory signal' such as human labels in data, STUM does not have an explicit `supervisory signal' and the supervision may come from any time-domain neighbors. Similar to unsupervised learning, STUM does not use labels for organizing data. Unlike traditional unsupervised learning which does not have learning guidance or supervision involvement across modalities, STUM gets learning guidance from time relation of data within and across modalities. Similar to reinforcement learning, time also plays an important role for the STUM model. Different from reinforcement learning which propagate scattered reward signals through time, STUM gets feedback from a function of time difference which is available at any moment.

Similar to GAN \cite{goodfellow2014generative}, both GAN and STUM have two units, one for generating outputs and another one for evaluating the performance. On the other hand, the focuses of GAN and STUM are different while traditional GAN demands high quality of generated data and STUM demands underlying meaning matches across modalities. Since these different focuses do not conflict with each other, it is possible to combine these demands in one optimization framework.

Deniz et al \cite{Deniz19} found that the representation of semantic information across human cerebral cortex during listening versus reading is invariant to stimulus modalities. More specifically, they find that reading a word or listening the same word stimulate the same location in the brain. The shared feature space of the STUM model aligns with this recent finding. It is different from deep neural network approaches that directly share hidden layers of different modalities \cite{Ngiam2011,Lu2015}. With direct hidden layer sharing, having an audio input only, a visual input only, or audiovisual combined inputs with the same semantic meaning may generate quite different vectors in the shared latent space. It means that the input of one modality may interfere the input representation of another modality. This arrangement may make the representation space chaotic, and make data organization of multiple modalities difficult. In the STUM model, the audio and visual channels will not interfere with each other. In other words, if audio and visual inputs have the same semantic meaning, they will activate the same vector space location no matter they are fed to the system separately or jointly. This property will make data organization for multiple modalities independent of modality count.

The STUM model is related to the recurrent visual attention model \cite{Gattention}. Similar to the recurrent visual attention model, STUM also has the assumption of narrow visual view based on research in \cite{Yarbus1967} and both approaches involve time in the learning. Different from the recurrent visual attention work that learns an attention prediction network based on current location and glimpse sensor info, STUM uses a simple time function to define the feature embedding distance and uses that to 'supervise' the formation of the feature space.

The STUM model is also related to autoencoders such as the Variational Autoencoder (VAE) \cite{KingmaW13} and the Adversarial Autoencoder (AAE) \cite{Goodfellow2016}. Similar to an autoencoder, the STUM architecture has both encoding the decoding units and a hidden layer connects its encoder and decoder. Different from a traditional autoencoder whose hidden layer is trained through back-propagation from decoder output to encoder input without time cue, the STUM hidden layer is organized through contrastive learning of time-windowed data and encoder training and decoder training are separated. This encoder and decoder training separation makes it easier for us to check independent encoder and decoder performance. Moreover, the STUM feature space clustering information is used to select representative output for feature space to output training. This output training sample selection process may overcome output blurry issues exist in traditional autoencoders, and give users more clear interpretation of the embedding space.

Vukotic et al. and Chaudhury et al \cite{Vukotic2016,Chaud2017} proposed models to learn cross-modality data transfer. They require data from both modalities for learning. The STUM model does not have that requirement for learning its feature space. Moreover, STUM uses guidance based on time-relation of data, which is not used in those approaches. Articles \cite{Harwath2016,Harwath2018} use trained dot-products of image and audio features to illustrate relations between objects in an image and words in a sentence. While it provides a great heat-map visualization in a compressed space-and-time domain, the relation indication is still weak and the network cannot use input in one modality to generate output in another modality like a human. Instead of generating outputs with the network, this approach requires the system to save both image and speech datasets for using data in one modality to retrieve the associated data in the opposite modality.

\section{Dataset}
Alan Turing suggested two approaches in his paper \cite{AMTuring} for a machine to achieve human like intelligence. His first suggestion is through a very abstract activity, like the playing of chess, which was heavily explored in the past 70 years. His second suggestion is to 'provide the machine with the best sense organs that money can buy, and then teach it to understand and speak English. This process could follow the normal teaching of a child. Things would be pointed out and named, etc.' Even though his second suggestion was barely explored in the past, it is an interesting path to overcome problems for learning from IoT sensors. Inspired by this suggestion, we choose image and speech as sensor inputs and use an object spoken name learning task for testing the STUM learning approach. For algorithm evaluation, we cannot use real-time training data in a random environment for unrepeatable tests. To overcome this problem, we have to create data that can simulate the eye and ear perceptions and can provide consistent inputs when we test different network options. It is better to compose this dataset with existing datasets so that we can have a better understanding of the training and testing effectiveness. The purpose of the new dataset is to simulate sensor captured data in real environment. The simulation is not part of the STUM learning algorithm.

To simulate the eye perception, we should understand that the size of the fovea (about 1–2 degrees of vision) is relatively small with regard to the rest of the retina, and it is the area where fine detail can be distinguished \cite{Jacobs2008}. Because of that, we assume that the perceived visual frames mainly focus on single object or a small portion of a `big' (depend on viewing distance) object. Moreover, eyes make short, rapid movements (saccades) intermingled with smooth pursuit \cite{Gegen2016}. Vision before and during saccades is very poor \cite{Yarbus1967,Volk1962}. So, saccades may be considered as start and stop positions of a visual time-window for learning/recognizing an object and the fixation duration may be considered as a learning time-window. Because of the eye movement pattern and narrow-field-of-view constraints, consecutive frames perceived by a human should mainly come from the same object before a saccade starts. Those consecutive frames may have small lighting changes or observation angle changes as we all experienced daily.

Since the proposed approach is quite different from existing approaches, there is no ready-to-use dataset for testing the approach. Because most well-known big datasets, such as ImageNet \cite{imagenet_cvpr09}, CIFAR10/CIFAR100 \cite{CIFAR09}, or MS COCO \cite{mscoco}, do not collect images in a continuous way, they cannot be used to simulate continuous visual inputs either. For example, the time duration of showing an object to a baby lasts several seconds to tens of seconds and the duration of a neuron excitation lasts tens to hundreds of milliseconds. To simulate those teaching scenarios, we have to find many sets of consecutive frames for various duration. The datasets mentioned earlier in this paragraph do not have images for this task. After examining many existing datasets, we found that the Columbia Object Image Library (COIL-100) \cite{Coil100} can meet most criteria for this simulation task.

The COIL-100 dataset contains color images of 100 objects, and these images were taken at pose intervals of 5 degrees from 0 to 360 degrees for each object. With these images, we can simulate what a human perceives when she/he looks through 100 objects. Even though a person may not see each object at every 5 degree angle, consecutive images captured from each object are very close to images a person perceived when she/he hold an object in hand and check the object. Moreover, compared with cluttered images, these COIL-100 images are more similar to what is captured by the human fovea (about 1-2 degrees of vision). This similarity can make the simulation closer to real perceptions. For audio segments corresponding to these objects, we create 100 English names, such as `mushroom', `Cetaphil' etc., and use Watson Text to Speech to generate 50 different audio segments for each object by varying voice model parameters such as expression etc. Overall, the new multimodal dataset has 72 consecutive images and 50 different audio segments for each object. With this new dataset, we randomly select 16 images from each object category as testing images and use the rest 56 images as training images. We also randomly select 10 audio segments from each object category and use the rest 40 audio segments in each category as training audios for our simulated data generation.

\begin{figure}[t]
\vskip 0.0in
\begin{center}
\centerline{\includegraphics[width=0.95\columnwidth] {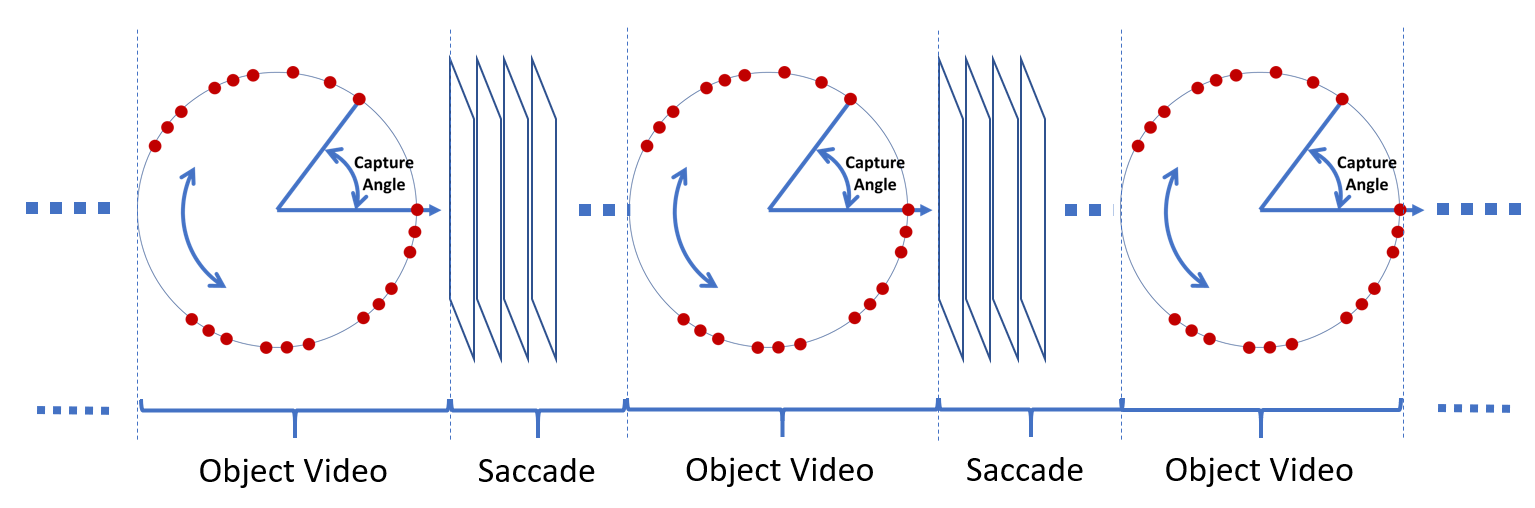}}
\caption{Mechanism for generating simulated video sequences. Each red dot on a circle represents an image captured at a certain angle. The system arranges randomly selected 56 images captured for each object in a circular mode based on capture angles. A video simulator will load images arranged with a ring pattern in a clockwise or counter clockwise order for simulating `smooth pursuit' video sequence focused on an object. The `smooth pursuit' video sequences are interleaved with blank `saccade' video sequences.
\label{fig:circular}}
\end{center}
\vskip -0.3in
\end{figure}

In our simulation, we assume that the learner's eye will focus on each object for certain amount of time before moving to a different object through a `saccade' period. With this assumption, the object `smooth pursuit' video sequences and blank `saccade' video sequences will be interleaved in the long generated sequence. To simulate an object video sequence perceived by a human eye, we arrange 56 training images in each category in a ring format based on their capture angles as that shown in Figure \ref{fig:circular}. Since we randomly put aside 16 images in each category for testing, the capture angle difference between any two consecutive training images in a ring is not always 5 degrees. That setting may simulate image angles missed by human observers or simulate variable speed when a person rotates an object in hand.

For generating a video sequence corresponding to the fixation (or smooth pursuit) period, the system can randomly select a starting point on a circle and pick a sequence of frames in clockwise or counter clockwise order. Because the images in each category have a cyclic order as that shown in Figure \ref{fig:circular}, the image picking process can proceed until the simulated fixation duration is reached. When a human moves eye from one object to another, the eye goes through a 'saccade' period. Since images captured during the 'saccade' period are not processed by human brain, we may use blank frames to simulate frames captured during a 'saccade' period. This video sequence generation mechanism is illustrated in Figure \ref{fig:circular}.

With a video generation mechanism ready, we need concrete parameters for generating a simulated video. Without losing generality, we choose the standard 30 frames/sec frame rate for video generation. For learning an object, we know that an ordinary teacher or student will not spend hours to look at the object. That learning duration is not in milliseconds range either. Based on our experience on teaching a baby, we choose random video length between 3 and 4 seconds for showing each object without losing generality. When a human moves eye from one object to another, the saccade normally take about 200 milliseconds to initiate, and then last about 20~200 ms \cite{saccadedur}. To simulate the saccade period, we insert 200ms-400ms blank frames between videos of objects.

\section{Time Cue Guided Learning}
The proposed STUM approach originates from our observation of the human visual system. Because of the inertia of natural objects and human eye, human's visual perception system may be viewed as a mechanical scanner which converts spatial relation of images to temporal relation of images. It may also be viewed as a physical nonlinear dimensionality reduction from the spatial domain to the time domain. More specifically, human's narrow fovea vision enables people to focus on single object or part of an object. Because of this narrow vision, human eye has to scan the space. During a scan, views that are close to each other in space (e.g. views from different angles of a small object) are usually close to each other in time when they get into the human's perception system. On the other hand, views that are close to each other in space are likely related to the same object. Therefore, we believe that time is an important cue for organizing object-related data within and across modalities. Inspired by the `fire together wire together' idea \cite{Hebb49} from biological science, we propose `connecting'/organizing features within a short time interval (i.e. tens of milliseconds to tens of seconds) together in the feature space to mimic the human learning process. We believe that this biological inspired approach can overcome many issues described earlier.

Similar to the normal teaching of a child, we train the system by showing objects and pronouncing their names at the same time. The proposed system is illustrated in Figure \ref{fig:archfig}. Audio
and image are two modalities in the illustration. A horizontal 'Time $t$' axis from left to right is used to reveal audio-visual changes over time. To make the audio operation simpler, we converted audio inputs to 2D logarithmic mel-spectrogram so that the system can handle audio in a similar way as it handles images.

According to the Hebbian theory in neural science, any two cells or groups of cells that are repeatedly active at the same time will tend to become `associated' so that activity in one facilitates activity in the other \cite{Hebb49}. Donald Hebb coined this finding as `Fire Together Wire Together'. In the STUM model, a shared feature space in the middle of Figure \ref{fig:archfig} is used to simulate the `Wire Together' process and a time window is used to define `Fire Together' which leads to `Wire Together' in the feature space. More specifically, if two or more feature vectors happen in the same time-window, the model will force them to get close to each other in the feature space. For feature vectors that happen outside of the time-window, the model will force them to stay away from features in the time-window to simulate the long-term depression (LTD) process which enables weakening of cell connections and eventually eliminates poor connections \cite{Massey2007,Anne2016}. This enforcement takes effect no matter the data are from the same modality or different modalities. With this mechanism, the model will form an embedding feature space shared by the vision and the speech modalities.

In Figure \ref{fig:archfig}, a time-window is marked with red dash line around some input images and an audio waveform. All inputs are transformed to the same feature space marked by a green dash line on the right of the image. Images in the time window are transformed to features by shared-weights visual CNN networks and their features attract each other (`wire together' near the red dot representation in the feature space) through a contrastive loss function. The red dot in the figure is an illustration for enforcing a representation cluster . It is not a fixed point in the feature space. To balance the feature space formation, inputs outside of the time-window are used as negative training samples. With the contrastive loss function, the negative sample feature (i.e. the blue dot in the feature space) and the anchor feature (the red dot) will repel each other. This repelling process may be considered as a process corresponding to the neural LTD process. With this mechanism, different appearances of the same object are clustered in the feature space while features of different objects will stay away from each other without label guidance.

Similar to the visual feature organization, the audio input within the time window is transformed to the same feature space through an audio CNN which is different from a visual CNN in the architecture. For audio input within the time-window, its representation and corresponding visual input representations in the same time-window  will attract each other (close to the red dot representation). The enforcement is also achieved through a contrastive loss function. Moreover, audio inputs outside of the time-window can be used as negative samples for the audio CNN training. During training, a negative audio sample feature (a blue dot) and features happened in the time-window will repel each other. For visual inputs that are far away from each other in space but have the same spoken name, the cross-modality feature organization process may help the system to cluster features of these inputs together.

\subsection{Models}
Learning features' time-relations such as 'Fire Together Wire Together' and the LTD process may be modeled with Equation \ref{eq:lossfunc4}:
\begin{equation}
	\mathit{H(t_i)} = \mathit{S(\| t_i - t_a\|)} ,
\label{eq:lossfunc4}
\end{equation}
where $H(t_i)$ is a sensory learning function which defines feature distance with time distance. The time distance only depends on the feature firing time $t_i$ and an anchor time $t_a$, and $t_a$ may be the 'present time' in an online learning system. With this learning function, we may further use Equations \ref{eq:lossfunc2} and \ref{eq:lossfunc6} to define the learning process:
\begin{equation}
	\mathit{D_{W_I,W_A}(\vv{X(t_i)},\vv{X(t_a)})} = \mathit{\big\| F(\vv{X(t_i)}) - F(\vv{X(t_a)})\big\|} ,
\label{eq:lossfunc2}
\end{equation}
\begin{equation}
	\mathit{L(W_I,W_A;\vv{X(t_i)})} = \mathit{\sum_{t_i}\big\|D(\vv{X(t_i)},\vv{X(t_a)})-H(t_i)\big\|}, \\
\label{eq:lossfunc6}
\end{equation}
where $X(t_i)$ is the $i^{th}$ sensory input, $D(t_i)$ is the  distance between the feature of input signal at time $t_i$ and input signal at time $t_a$, and $F(t_i)$ and $F(t_a)$ are the $i^{th}$ feature representation and anchor feature representation respectively. These features are high-dimensional outputs of the corresponding conversion networks with feature encoding weights $W_I$ and $W_A$ for the image CNN channel and the audio CNN channel respectively. The loss function $L$ in Equation \ref{eq:lossfunc6} enforces the system to form feature space through aligning feature distances with expected feature distances (i.e. the sensory learning function). For a simple time-window based approach, we may simply define $H(t_i)$ to be 0 within the window close to an anchor time for simulating the wiring together process and 1 outside of window for simulating the long-term depression (LTD) process. By enforcing $t_i<t_a$, the system can perform online learning with streaming data.

For a simple window-based simulation, above learning equations may be simplified with a contrastive loss function similar to the paper \cite{LeCun2006}. Corresponding to the illustration in Figure \ref{fig:archfig}, the loss function can be described with Equation \ref{eq:lossfunc1}:
\begin{equation}
    \begin{multlined}
	\mathit{L_{W_I,W_A}(\vv{X(t_i)})} = \mathit{\sum_{i}\Big\{ \big(1 - Z(t_i)\big)D(t_i)} \\
	+ \mathit{Z(t_i)\big[max(0,m-D(t_i))\big]\Big\}},
	\end{multlined}
\label{eq:lossfunc1}
\end{equation}
where $Z(t_i)$ is a `Fire Together' indicator which is 0 for media happening within the time-window and 1 for media  happening outside the time window, $m$ is a fixed margin in the shared feature space.

\subsection{Learning for Embedding Feature Interpretation}
To understand the state of a learning system, it is better that the learning system can interpret its feature space activation with a human recognizable media. If inputs to the learning system are common media that are also available to human, interpreting the feature space activation in those media format is an easy way to communicate with human. For example, when a teacher wants to check the progress of a student on naming an object, the teacher may show an object to the student for student to give a proper pronunciation. The teacher may also pronounce the name of an object for the student to find the real object or draw the object. For the system to learn internal representation to output media mapping, the system may automatically pick the window-selected input whose feature is the closest to the mean of all features that happen in the same time-window, and use it as a training sample for the output. This selection process can be done with:

\begin{equation}
	\mathit{\vv{Y}}=\mathit{\Big\{\vv{X_{i^*}} \;\big|\; i^* = \argminA_{i \in I} {\big\| \vv{F_i} - \vv{F_{av}}\big\|}\Big\}} ,
\label{eq:lossfunc3}
\end{equation}
where $\vv{Y}$ is the selected training output, $I$ is the input index set selected by the time-window and $\vv{F_{av}}$ is the average of all input features chosen by the time-window. If the inverse of the distance between a feature and the average feature is related to the strength of cell activation, the selected input is the one that has the strongest activation. By selecting a representative output for a feature cluster, the system will have more chances to get a clear output for each cluster. With feature representations and corresponding output training samples, audiovisual output networks can be trained.

\section{Evaluations}
\subsection{Time Cue guided Visual Learning}
Since humans can learn object visual concepts without audio, we first test our approach with vision modality only. In other words, we only generate the training video sequence with the COIL-100 set and the aforementioned generation mechanism. For the learning process, we use Equation \ref{eq:lossfunc1} and Equation \ref{eq:lossfunc2} without the optimization of $\mathit{W_A}$. In the evaluation network, the image encoding module has 7 convolution layers, a $128 \times 128$ $3$-channel color input, and a 1024-dimension feature space. It also has matched down sampling layers, normalization layers, and Leaky ReLU layers.

\begin{figure}[t]
    \centering
    {\includegraphics[width=0.45\columnwidth] {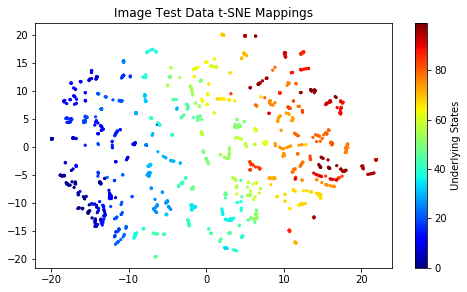}}
    {\includegraphics[width=0.45\columnwidth] {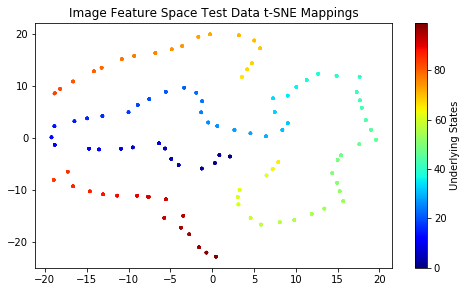}}
    \caption{Input space and feature space comparison for the network trained with visual modality only. 100 different colors represent 100 different underlying states corresponding to different objects. Left: Testing image input t-SNE mapping (1600 samples without clear clusters). Right: Feature space t-SNE mapping corresponding to these inputs. There are 100 clear clusters in this map.}
    \label{fig:tsne1modal}
    \vspace*{\floatsep}
\vskip -0.2in
\end{figure}

To simplify the training, we pick two consecutive image frames as a 'Fire Together' pair corresponding to the first term in \ref{eq:lossfunc1}, and pick the frame that is 300 frames away from the focused frame as a negative pair corresponding to the second term in Equation \ref{eq:lossfunc1}. In other words, we assume that people do not stare at one spot for over 10 seconds without losing generality. For online learning, the anchor visual frame can be the last visual frame input, its 'Fire Together' pair is the second last frame, and its negative pair is a past frame 300 frames away. Corresponding to Equation \ref{eq:lossfunc4} has value 0 near the current time, value 1 at the time 300 frames away, and no definition for other positions. For batch processing, 56,000 pairs are selected in this way for training the learning network. We then feed 1,600 testing images to the network to compute their features. Figure \ref{fig:tsne1modal} shows the comparison between the input samples and their features. On the left of this figure, we can see the 1,600-input-sample distribution which does not have clear clusters. On the right of this figure, we can see 100 clear clusters in the feature space. These clusters unveil clear concepts learned by the network.

In the 1,600 test images, there are 100 categories and each category has 16 randomly selected images. To get quantitative measurement on feature clustering, we randomly pick an image from the same category for each of these 1,600 test images to form 1,600 positive pairs, and randomly pick an image from a different category for each test image to form negative pairs. We then feed these 3,200 pairs to the network and calculate the feature distance for each pair. With a simple threshold half way between the maximum distance and minimum distance, we get 99.375\% accuracy on predicting if a testing pair belongs to the same category or not. This high prediction accuracy further convinced us the good clustering performance.

\subsection{Time Cue guided Multi-modality Learning}
Beyond learning concepts from single modality data, human can also learn data relation across modalities. For example, human can learn the spoken name of an object by pronouncing the spoken name while showing the object. To simulate the naming object training scenarios, we create a signal generation machine to mimic a human teacher (it may also mimic a natural multimedia bounding that is enforced by the environment). In the evaluation experiment, the signal generation machine uses above training dataset (i.e., 56 images and 40 audio segments in each category for 100 categories) and the training video generation mechanism to generate synchronized image sequences and corresponding audios. For the naming object training task, the system sets 100 underlying states corresponding to 100 object categories for generating co-occurred image sequences and audio signals. These underlying states are unavailable to the learning machine so that we can check if related concepts can be formed through learning from sensors. With this setup, when the training machine wants to teach a concept (e.g. `Advil'), it generates the `Advil' spoken name audio with a sequence of `Advil' consecutive images and forward them `together' to the learning system. This is similar to the teaching process when a teacher shows the `Advil' object and pronounce `Advil' at the same time.

To make the speech-vision synchronization easier, we convert all audio wave-forms to logarithmic mel-spectrograms. More specifically, we use audio tools by Kastner \cite{Kastner17} to perform the conversion. The logarithmic mel-spectrogram transformation has 64 Mel filter bands, a 46.4ms transformation window, and a 33ms shift step. With the audio-window time step equal to video frame time difference, it is easy to align audio data with a video sequence. Since speech waveforms have variable lengths, the 5,000 spectrograms have different number of 64-dimension vectors. As we described in early paragraphs, we allocated 40 spectrograms in each category for the training system and 10 spectrograms in each category for testing use.

For synchronizing speech data with a generated single modality video sequence in a combined sequence, the system needs to find the middle frame index of an object video segment, find the middle vector index of a selected spectrogram, and align these two index positions. Since video sequences and audio sequences have variable lengths, selected spectrograms can be fit in a large spectrogram whose width equals to the number of frames of the long video sequence. Since the video sequence for each object is around 3-4 seconds (mimic the time period of a regular teaching process) and the longest speech in the dataset is around 1.7 seconds, it is guaranteed that these selected spectrograms have no overlap in the big spectrogram. For speech channel positions that are not occupied by aligned spectrogram data, the system will fill them with empty vectors. When the system's learning algorithm picks a video frame, the system can get its corresponding audio spectrograms by searching for high-energy sepctrogram section around +/- 2 seconds. With a 1.8s spectrogram time window, the system can get a 2D spectrogram image associated with the video frame. In this way, each video frame in the combined sequence has an audio spectrogram image associated with it. Similarly, a negative pair can be retrieved 300 frames away.

In the multi-modality joint learning network, the visual learning network has the same structure as the network used for isolated visual modality learning. The audio learning module has 6 convolution layers, a $64 \times 64$ 1-channel logarithmic mel-spectrogram input, and a 1024-dimension feature space. It also has matched down sampling layers, normalization layers, and Leaky ReLU layers. Both visual learning module and audio learning module share the same 1024-dimension feature space.

To simplify the training based on Equation \ref{eq:lossfunc1} and \ref{eq:lossfunc2}, the system forms a positive image pair with consecutive video frames and an audiovisual positive pair with a focused video frame and its associated 2D audio spectrogram.
The distortion caused by the positive pair corresponds to the first term in Equation \ref{eq:lossfunc1}. Similarly, the system forms a negative image pair with a focused video frame and a frame 300 frames away from the focused frame and an audiovisual negative pair with the focused video frame and the 2D audio spectrogram 300 frames away. The distortion caused by the negative pair corresponds to the second term in Equation \ref{eq:lossfunc1}. For batch processing, the system formed 56,000 image pairs and 56,000 audiovisual pairs to train the system.

\begin{figure}[t]
    \centering
    {\includegraphics[width=0.45\columnwidth] {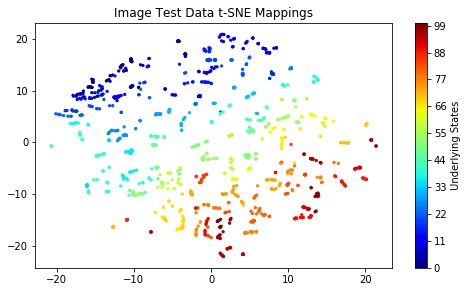}}
    {\includegraphics[width=0.45\columnwidth] {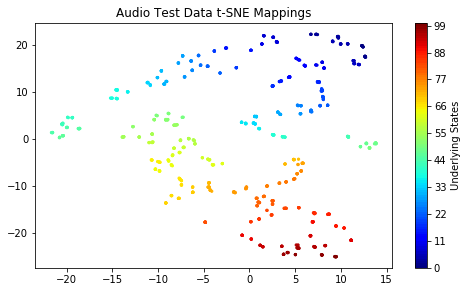}}
    {\includegraphics[width=0.9\columnwidth] {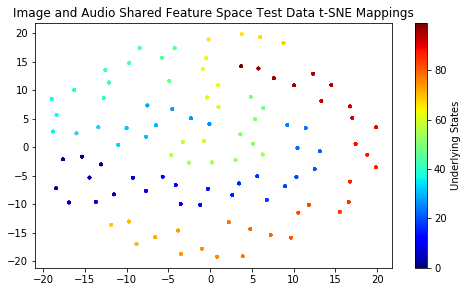}}
    \caption{Input space and feature space comparison for the network trained with audio and visual modalities. 100 different colors represent 100 different underlying states corresponding to different objects. Top Left: Testing image input t-SNE map (1600 samples). Top Right: Testing audio logarithmic mel-spectrogram input t-SNE map (1000 samples). Bottom: Audiovisual shared feature space t-SNE map (2600 mixed features in 100 clear clusters).}
    \label{fig:tsne}
    \vspace*{\floatsep}
\vskip -0.2in
\end{figure}

After the training, we feed 1,600 testing images and 1,000 testing audio spectrograms to the network to compute their features. To understand the training effect, we show t-SNE maps of audio and visual inputs, and their corresponding features in shared feature space in Figure \ref{fig:tsne}. In these maps, 100 different colors represent 100 different underlying concepts. From color distributions in these maps, we can see that data with the same underlying state form much more distinctive groups in the shared feature space than the two input spaces. This may be related to a `concept' formation process in the network. We believe these separated clusters may facilitate higher-level learning process in the future. In the feature space, the margins among groups of different states also increase significantly. The increased margins in the shared feature space and output space can help the system to reduce semantic mapping errors. It may also help the machine to reduce its communication error with humans.

Similar to the quantitative testing of the network trained with visual only input, we created both visual testing pairs and audiovisual testing pairs. The 3,200 visual testing pairs are created in the same way as we did in the previous section. Moreover, we randomly pick an audio spectrogram in the same category to form a positive pair, and randomly pick an audio spectrogram from a different category to form a negative pair for each testing image. In this way, we can get 3,200 audiovisual testing pairs. We feed these 6,400 testing pairs to the trained network and get maximum feature distance and minimum feature distance for all pairs. By using halfway between the minimum feature distance and maximum distance as a threshold, we can get 99.844\% accuracy on predicting if a visual testing pair belongs to the same category or not, and 100\% accuracy on predicting if an audiovisual testing pair belong to the same category or not. Both these results are better than the performance we get from single modality training. This performance improvement may be attributed to the cross-modality influence. More specifically, signal co-occurrences across modalities give the system more cues to cluster data in the feature space.  

\subsection{Input/Output Underlying-state Matching Performance}

\begin{figure}[t]
    \centering
    {\includegraphics[width=0.45\columnwidth] {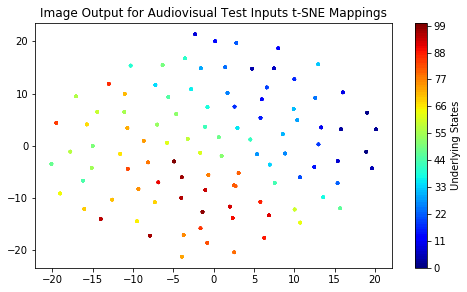}}
    {\includegraphics[width=0.45\columnwidth] {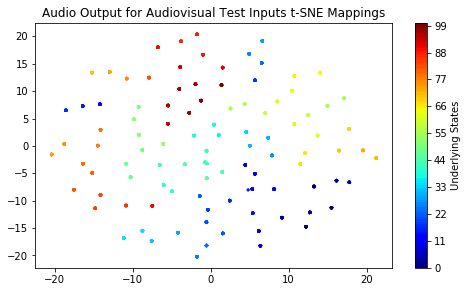}}
    \caption{Image and audio output signal t-SNE maps. The inputs include 1600 test images and 1000 test audios. 100 different colors represent 100 different underlying states corresponding to different objects. Left: Image output t-SNE map for both audio and visual inputs. Right: Audio output t-SNE map for both audio and visual inputs.}
    \label{fig:tsneoutput}
    \vspace*{\floatsep}
\vskip -0.2in
\end{figure}

When a teacher checks a student's object naming performance, the teacher may show an object to the student and asks the student to pronounce its name, or may pronounce the name of an object and asks the student to select the object or draw the object. Then, the teacher will grade the student based on the percentage of correct answers. Following the encoder training described above is the feature space to output training guided by Equation \ref{eq:lossfunc3}. Different from traditional autoencoders, STUM uses data-driven clusters formed in the feature space to assist the decoder training. More specifically, it collects features `stimulate' (i.e. reach) each cluster within the cluster threshold in a time window, computes the mean of these features, finds the input whose feature is the closest to the cluster mean, and uses the found input as output to train the mapping from those cluster features to the output. The process aligns with the scenario where people find the most 'standard' pronunciation or writing to mimic. In this process, the output training sample selection gives the system more chances to output high-quality image and speech. Figure \ref{fig:tsneoutput} shows audiovisual output t-SNE maps corresponding to those inputs shown in \ref{fig:tsne}. From the map, we can see 100 clear clusters. The overall network learning performance measurement follows a similar approach used by a human teacher. More specifically, we measure underlying states matching percentage of the network's encoder inputs and decoder outputs (e.g. the underlying states of $\bf{x_I}$ and $\bf{y_A}$).

\begin{table}[t]
\begin{center}
\caption{Comparison between STUM media conversions and text label supervised classifiers. Bottom four lines are results from testing the STUM system} \label{tab:comparison}
\vskip 0.1 in
\begin{tabular}{|c|l|r|}
  \hline
  Test ID   &Approaches                 & Acc($\%$)
  \\
  \hline
  &\textit{Category predicting accuracy with} & $ $ \\
  $1$ &\textit{network trained with visual data only} & $99.375$ \\
    \hline
  &\textit{Category predicting accuracy with} & $ $ \\
  $2$ &\textit{network trained with audiovisual data} & $99.844$ \\
  \hline
  $3$ &\textit{Image Label Classifier (ILC)}  & $100.00$  \\
    \hline
  $4$ &\textit{Audio Label Classifier (ALC)}  & $100.00$  \\
  \hline
  $5$ &\textit{COIL Image} $\mathit{ \rightarrow E_I \rightarrow D_A \rightarrow ALC}$   & $100.00$ \\
    \hline
  $6$ &\textit{TTS Speech} $\mathit{ \rightarrow E_A \rightarrow D_I \rightarrow ILC}$  & $100.00$ \\
    \hline
  $7$ &\textit{COIL Image} $\mathit{ \rightarrow E_I \rightarrow D_I \rightarrow ILC}$   & $100.00$ \\
    \hline
  $8$ &\textit{TTS Speech} $\mathit{ \rightarrow E_A \rightarrow D_A \rightarrow ALC}$  & $100.00$ \\
  \hline
\end{tabular}
\end{center}
\vskip -0.2in
\end{table}

To save human grading labor (i.e., checking input-output underlying-state matches for 1,000 testing audios and 1,600 testing images) and make the evaluation process more objective, we trained an audio pattern classifier and an image pattern classifier with labeled image and audio training data. The image classifier has a $99.81 \%\ $ labeling accuracy on 1,600 testing images and the audio classifier has a $100 \%\ $ labeling accuracy on 1,000 audio testing segments. These data are shown in Table \ref{tab:comparison}. Due to high accuracies on the test data, we believe the trained classifiers are reliable to judge the underlying-state matching correctly.

After classifiers' training, we feed 1,600 testing images to the system to get 1,600 audio outputs and 1,600 image outputs , and these image and audio outputs are fed to the corresponding image or audio classifier to predict the underlying states. Similarly, we feed 1,000 testing audio segments to the system to get 1,000 image outputs and 1,000 audio outputs. These image and audio outputs are then fed to the corresponding image or audio classifier for predicting their underlying states. The matching percentages of the input-output underlying states are reported in the Table \ref{tab:comparison}. In the table, the $5^{th}$ experiment shows the performance of `naming' a COIL test image. The $6^{th}$ experiment shows the performance of `imagining'/`drawing' an image based on a testing speech input. The $7^{th}$ experiment shows the performance of seeing an object image and `drawing' a representative image. The $8^{th}$ experiment is the performance of hearing a spoken name and repeat the speech. With random network parameters before the training process, the accuracy numbers corresponding to the setups of last four lines are around ${1 \%}$. After learning from the training signals, the accuracy in the last four lines of the table can all reach perfect without using any labels. These results further convinced us the effectiveness of this no-text-label time-cue guided learning approach.

\section{Conclusion and Future Work}
In this paper, we propose STUM for learning from multi-modal sensor inputs. Contributions of the paper include a time-cue guided sensor signal learning model within and across modalities, an eye movement inspired signal co-occurrence model, an audiovisual dataset aligned with the model, and a performance evaluation method that mimics a teacher's grading process. After training, the STUM networks can convert signal in one modality to signal in another modality or convert the input signal to a more representative signal. The high performance achieved by the STUM system convinces us the feasibility of machine learning without text labels. Furthermore, the training can be achieved without future data. That makes it possible for online learning with the STUM model. Beyond self-contained learning, we may also consider using the STUM system before a traditional image recognizer to improve the overall recognizing accuracy on objects. Additionally, media association may be helpful for joint visual-speech recognition.

Through experiments, we found clear data clusters correlated with underlying states. We believe these separated clusters may facilitate higher-level `concept' learning process. Good evaluation results further convinced us the idea of learning directly from sensory data. It also shows us research `head-space' for creating new dataset, new learning model, new training approaches, and new applications. Even though we only use data of isolated objects in this experiment, we think this method may be extended to continuous scenes that don't have a clear boundary. For example, associate a group of scenes with certain segment or location of a road.  We think further research in this direction may enable us to provide `eyes' and `ears' etc. to an autonomous computer and program the computer in a way that a teacher instructs a student. It also has potential to greatly reduce offline text labeling. In the future, we plan to test the system in a more sophisticated scenario.


%

\ifCLASSOPTIONcaptionsoff
  \newpage
\fi



%



\bibliographystyle{IEEEtran}
\bibliography{Stum}

\begin{thebibliography}{10}
\providecommand{\url}[1]{#1}
\csname url@samestyle\endcsname
\providecommand{\newblock}{\relax}
\providecommand{\bibinfo}[2]{#2}
\providecommand{\BIBentrySTDinterwordspacing}{\spaceskip=0pt\relax}
\providecommand{\BIBentryALTinterwordstretchfactor}{4}
\providecommand{\BIBentryALTinterwordspacing}{\spaceskip=\fontdimen2\font plus
\BIBentryALTinterwordstretchfactor\fontdimen3\font minus
  \fontdimen4\font\relax}
\providecommand{\BIBforeignlanguage}[2]{{%
\expandafter\ifx\csname l@#1\endcsname\relax
\typeout{** WARNING: IEEEtran.bst: No hyphenation pattern has been}%
\typeout{** loaded for the language `#1'. Using the pattern for}%
\typeout{** the default language instead.}%
\else
\language=\csname l@#1\endcsname
\fi
#2}}
\providecommand{\BIBdecl}{\relax}
\BIBdecl

\bibitem{Pete2018}
P.~Warden, ``Why the future of machine learning is tiny,''
  https://petewarden.com/2018/06/11/why-the-future-of-machine-learning-is-tiny/.

\bibitem{Thrun1998}
S.~Thrun and L.~Pratt, Eds., \emph{Learning to Learn}.\hskip 1em plus 0.5em
  minus 0.4em\relax USA: Kluwer Academic Publishers, 1998.

\bibitem{Vinyals2016}
O.~Vinyals, C.~Blundell, T.~Lillicrap, K.~Kavukcuoglu, and D.~Wierstra,
  ``Matching networks for one shot learning,'' in \emph{Proceedings of the 30th
  International Conference on Neural Information Processing Systems}, ser.
  NIPS'16.\hskip 1em plus 0.5em minus 0.4em\relax Red Hook, NY, USA: Curran
  Associates Inc., 2016, p. 3637–3645.

\bibitem{Snell2017}
\BIBentryALTinterwordspacing
J.~Snell, K.~Swersky, and R.~Zemel, ``Prototypical networks for few-shot
  learning,'' in \emph{Advances in Neural Information Processing Systems 30},
  I.~Guyon, U.~V. Luxburg, S.~Bengio, H.~Wallach, R.~Fergus, S.~Vishwanathan,
  and R.~Garnett, Eds.\hskip 1em plus 0.5em minus 0.4em\relax Curran
  Associates, Inc., 2017, pp. 4077--4087. [Online]. Available:
  \url{http://papers.nips.cc/paper/6996-prototypical-networks-for-few-shot-learning.pdf}
\BIBentrySTDinterwordspacing

\bibitem{Sung2018}
F.~{Sung}, Y.~{Yang}, L.~{Zhang}, T.~{Xiang}, P.~H.~S. {Torr}, and T.~M.
  {Hospedales}, ``Learning to compare: Relation network for few-shot
  learning,'' in \emph{2018 IEEE/CVF Conference on Computer Vision and Pattern
  Recognition}, 2018, pp. 1199--1208.

\bibitem{LeCun2006}
R.~Hadsell, S.~Chopra, and Y.~LeCun, ``Dimensionality reduction by learning an
  invariant mapping,'' in \emph{In Proceedings of the 2006 IEEE Computer
  Society Conference on Computer Vision and Pattern Recognition}, C.~Society,
  Ed., vol.~2.\hskip 1em plus 0.5em minus 0.4em\relax Washington, DC, USA:
  IEEE, 2006, pp. 1735--1742.

\bibitem{Baldi1993}
P.~Baldi and Y.~Chauvin, ``Neural networks for fingerprint recognition,''
  \emph{Neural Computation}, vol.~5, 05 1993.

\bibitem{longuet1981}
H.~C. Longuet-Higgins, ``A computer algorithm for reconstructing a scene from
  two projections,'' \emph{Nature}, vol. 293, no. 5828, pp. 133--135, 1981.

\bibitem{Hartley2004}
R.~I. Hartley and A.~Zisserman, \emph{Multiple View Geometry in Computer
  Vision}, 2nd~ed.\hskip 1em plus 0.5em minus 0.4em\relax Cambridge University
  Press, ISBN: 0521540518, 2004.

\bibitem{Hoffer2015}
E.~Hoffer and N.~Ailon, ``Deep metric learning using triplet network,''
  \emph{Similarity-Based Pattern Recognition. SIMBAD 2015. Lecture Notes in
  Computer Science}, vol. 9370, 2015.

\bibitem{Bengio2009}
G.~Chechik, V.~Sharma, U.~Shalit, and S.~Bengio, ``Large scale online learning
  of image similarity through ranking,'' \emph{Pattern Recognition and Image
  Analysis}, vol. 5524, 2009.

\bibitem{goodfellow2014generative}
I.~Goodfellow, J.~Pouget-Abadie, M.~Mirza, B.~Xu, D.~Warde-Farley, S.~Ozair,
  A.~Courville, and Y.~Bengio, ``Generative adversarial nets,'' in
  \emph{Advances in neural information processing systems}, 2014, pp.
  2672--2680.

\bibitem{Deniz19}
\BIBentryALTinterwordspacing
F.~Deniz, A.~O. Nunez-Elizalde, A.~G. Huth, and J.~L. Gallant, ``The
  representation of semantic information across human cerebral cortex during
  listening versus reading is invariant to stimulus modality,'' \emph{Journal
  of Neuroscience}, 2019. [Online]. Available:
  \url{https://www.jneurosci.org/content/early/2019/08/16/JNEUROSCI.0675-19.2019}
\BIBentrySTDinterwordspacing

\bibitem{Ngiam2011}
J.~Ngiam, A.~Khosla, M.~Kim, J.~Nam, H.~Lee, and A.~Ng, ``Multimodal deep
  learning,'' in \emph{Proceedings of International Conference on Machine
  Learning 2011}, Bellevue, Washington, USA, 2011.

\bibitem{Lu2015}
H.~Lu, Y.~Liou, H.~Lee, and L.~Lee, ``Semantic retrieval of personal photos
  using a deep autoencoder fusing visual features with speech annotations
  represented as word/paragraph vectors,'' in \emph{Proceedings of Annual
  Conference of the International Speech Communication Association}, 2015.

\bibitem{Gattention}
\BIBentryALTinterwordspacing
V.~Mnih, N.~Heess, A.~Graves, and K.~Kavukcuoglu, ``Recurrent models of visual
  attention,'' \emph{CoRR}, vol. abs/1406.6247, 2014. [Online]. Available:
  \url{http://arxiv.org/abs/1406.6247}
\BIBentrySTDinterwordspacing

\bibitem{Yarbus1967}
A.~L. Yarbus, \emph{Eye Movements and Vision}.\hskip 1em plus 0.5em minus
  0.4em\relax Plenum. New York., 1967.

\bibitem{KingmaW13}
D.~P. Kingma and M.~Welling, ``Auto-encoding variational bayes,'' \emph{CoRR},
  vol. abs/1312.6114, 2013.

\bibitem{Goodfellow2016}
\BIBentryALTinterwordspacing
A.~Makhzani, J.~Shlens, N.~Jaitly, and I.~Goodfellow, ``Adversarial
  autoencoders,'' in \emph{International Conference on Learning
  Representations}, 2016. [Online]. Available:
  \url{http://arxiv.org/abs/1511.05644}
\BIBentrySTDinterwordspacing

\bibitem{Vukotic2016}
V.~Vukotic, C.~Raymond, and G.~Gravier, ``Bidirectional joint representation
  learning with symmetrical deep neural networks for multimodal and crossmodal
  applications,'' in \emph{Proceedings of ACM International Conference on
  Multimedia Retrieval}, New York, United States, 2016.

\bibitem{Chaud2017}
S.~Chaudhury, S.~Dasgupta, A.~Munawar, M.~A.~S. Khan, and R.~Tachibana,
  ``Conditional generation of multi-modal data using constrained embedding
  space mapping,'' in \emph{Proceedings of ICML Workshop on Implicit Models},
  Sydney, Australia, 2017.

\bibitem{Harwath2016}
D.~Harwath, A.~Torralba, and J.~Glass, ``Unsupervised learning of spoken
  language with visual context,'' in \emph{Proceedings of Neural Information
  Processing Systems (NIPS) (2016)}, Barcelona, Spain, 2016.

\bibitem{Harwath2018}
D.~Harwath, A.~Recasens, D.~Surıs, G.~Chuang, A.~Torralba, and J.~Glass,
  ``Jointly discovering visual objects and spoken words from raw sensory
  input,'' in \emph{Proceedings of European Conference on Computer Vision
  (ECCV) 2018}, Munich, Germany, 2018.

\bibitem{AMTuring}
\BIBentryALTinterwordspacing
A.~M. TURING, ``{I.—COMPUTING MACHINERY AND INTELLIGENCE},'' \emph{Mind},
  vol. LIX, no. 236, pp. 433--460, 10 1950. [Online]. Available:
  \url{https://doi.org/10.1093/mind/LIX.236.433}
\BIBentrySTDinterwordspacing

\bibitem{Jacobs2008}
G.~H. Jacobs, ``Primate color vision: a comparative perspective,'' \emph{Vis
  Neurosci}, vol. 25(5-6), p. 619–33, 2008.

\bibitem{Gegen2016}
K.~R. Gegenfurtner, ``The interaction between vision and eye movements,''
  \emph{Perception}, vol. 45 (12), p. 1333–1357, 2016.

\bibitem{Volk1962}
F.~C. Volkmann, ``Vision during voluntary saccadic eye movements,''
  \emph{Journal of the Optocal Society A}, vol.~52, p. 571–577, 1962.

\bibitem{imagenet_cvpr09}
J.~Deng, W.~Dong, R.~Socher, L.-J. Li, K.~Li, and L.~Fei-Fei, ``{ImageNet: A
  Large-Scale Hierarchical Image Database},'' in \emph{CVPR09}, 2009.

\bibitem{CIFAR09}
A.~Krizhevsky, ``Learning multiple layers of features from tiny images,''
  Canadian Institute for Advanced Research, Tech. Rep., 2009.

\bibitem{mscoco}
\BIBentryALTinterwordspacing
T.~Lin, M.~Maire, S.~J. Belongie, L.~D. Bourdev, R.~B. Girshick, J.~Hays,
  P.~Perona, D.~Ramanan, P.~Doll{\'{a}}r, and C.~L. Zitnick, ``Microsoft
  {COCO:} common objects in context,'' \emph{CoRR}, vol. abs/1405.0312, 2014.
  [Online]. Available: \url{http://arxiv.org/abs/1405.0312}
\BIBentrySTDinterwordspacing

\bibitem{Coil100}
S.~A. Nene, S.~K. Nayar, H.~Murase \emph{et~al.}, ``Columbia object image
  library (coil-100),'' Columbia University, Tech. Rep., 1996.

\bibitem{saccadedur}
H.~Kirchner and S.~Thorpe, ``Ultra-rapid object detection with saccadic eye
  movements: Visual processing speed revisited,'' \emph{Vision research},
  vol.~46, pp. 1762--76, 06 2006.

\bibitem{Hebb49}
D.~Hebb, \emph{The Organization of Behavior}.\hskip 1em plus 0.5em minus
  0.4em\relax New York: Wiley \& Sons, 1949.

\bibitem{Massey2007}
P.~Massey and Z.~Bashir, ``Long-term depression: multiple forms and
  implications for brain function,'' \emph{Trends Neurosci.}, vol. 30 (4), p.
  176–84, Apr. 2007.

\bibitem{Anne2016}
A.~Trafton, ``How neurons lose their connections,'' Jan. 2016, mIT News.

\bibitem{Kastner17}
\BIBentryALTinterwordspacing
K.~Kastner, ``Audio tools for numpy/python,'' \emph{GitHub repository}, 2018.
  [Online]. Available: \url{https://gist.github.com/kastnerkyle}
\BIBentrySTDinterwordspacing

\end{thebibliography}

%

\begin{IEEEbiography}[{\includegraphics[width=1in,height=1.25in,clip,keepaspectratio]{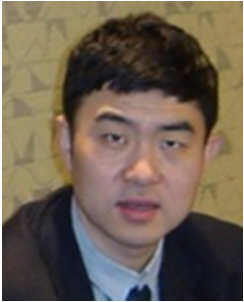}}]{Qiong Liu}
Qiong Liu is a principal scientist with Fuji-Xerox Palo Alto Laboratory (FXPAL). He got his Ph.D. from University of Illinois at Urbana-Champaign, his Master degree from Tsinghua University, and his Bachelor degree from Zhejiang University. He has served as the General Chair of ACM Multimedia 2017, the ACM SIGMM Best Ph.D. Thesis Committee (2018, 2019), NSF review panels, and Ph.D. Thesis Committee for UC Irvine and UC Santa Barbara.

Qiong currently has over 80 publications and 100+ patents (issued and pending) in multiple nations. He was nominated for best paper award four times in top ACM/IEEE conferences, receiving best paper award as a first author from ACM IUI 2010, and was nominated for best demos twice, receiving a best demo award from IEEE ICME 2014. He was also granted the Fuji-Xerox Technology Award by President Toshio Arima.

\end{IEEEbiography}

\begin{IEEEbiography}[{\includegraphics[width=1in,height=1.25in,clip,keepaspectratio]{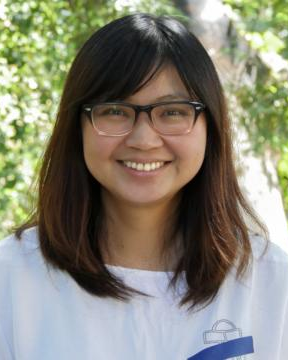}}]{Yanxia Zhang}
Yanxia Zhang is a research scientist at FX Palo Alto Laboratory (FXPAL). Her research focuses on human behavior sensing and analysis. Prior to her position at FXPAL, she was a post-doc researcher in the Pattern Recognition \& Bioinformatics Group in TU Delft, and a research fellow at the Royal Institute of Technology (KTH) in Sweden, in the Computer Vision and Active Perception Lab. She received her PhD degree in Computer Science from Lancaster University, obtained her Master’s degree in Artificial Intelligence from the University of Amsterdam, where she was awarded the Huygens Scholarship.
\end{IEEEbiography}






\end{document}